\documentclass{article}
\usepackage{spconf,amsmath,graphicx}
\usepackage{amsmath,amssymb,amsfonts}
\usepackage{algorithmic}
\usepackage{graphicx}
\usepackage{textcomp}
\usepackage{float}
\usepackage{xcolor}
\usepackage{booktabs} 
\usepackage{multirow}
\usepackage{makeidx}
\usepackage{enumitem}
\setlist{nosep, leftmargin=8pt}

\usepackage{mwe} 

\title{Parkinson's Disease Classification Using Contrastive Graph Cross-View Learning with Multimodal Fusion of SPECT Images and Clinical Features}
\name{Jun-En Ding $^{\star}$ \qquad  Chien-Chin Hsu $^{\dagger}$ \qquad  Feng Liu $^{\star}$ }
\address{$^{\star}$ School of Systems and Enterprises, Stevens Institute of Technology, Hoboken, NJ 07030, USA  \\ $^{\dagger}$ Dept. Nuclear Medicine, Kaohsiung Chang Gung Memorial Hospital, Kaohsiung Chang, Taiwan}

\begin{document}
%
\maketitle

\begin{abstract}
Parkinson's Disease (PD) affects millions globally, impacting movement. Prior research utilized deep learning for PD prediction, primarily focusing on medical images, neglecting the data's underlying manifold structure. This work proposes a multimodal approach encompassing both image and non-image features, leveraging contrastive cross-view graph fusion for PD classification. We introduce a novel multimodal co-attention module, integrating embeddings from separate graph views derived from low-dimensional representations of images and clinical features. This enables more robust and structured feature extraction for improved multi-view data analysis. Additionally, a simplified contrastive loss-based fusion method is devised to enhance cross-view fusion learning. Our graph-view multimodal approach achieves an accuracy of 0.91 and an area under the receiver operating characteristic curve (AUC) of 0.93 in five-fold cross-validation. It also demonstrates superior predictive capabilities on non-image data compared to solely machine learning-based methods.
\end{abstract}

\section{INTRODUCTION}
In recent years, deep learning models have shown great successes in biomedical applications, including computer-aided detection/diagnosis, image segmentation, image generation, disease staging and prediction,  based on the radiological images or physiological recordings, such as computed tomography (CT), Positron Emission Tomography (PET),  Magnetic Resonance Imaging (MRI) and electroencephalogram/Magnetoencephalography  (EEG/MEG)~\cite{shen2017deep,oh2020deep,jiao2022graph}. Despite their many strengths, convolutional neural network (CNN) models which is good at extracting high level features can be benefited by   incorporating all clinical factors at the same time for the downstream classification tasks. Some researches have focused on integrating medical image and clinical features with deep learning frameworks, which improves precision and the individualization of preventive, diagnostic, and therapeutic strategies ~\cite{acosta2022multimodal}. 

Graph representation learning that can incorporate node features  as well as utilize the connectivity/similarity information among all the nodes achieved good success for machine learning tasks on the graph structured data, such as graph convolutional neural network (GCN) ~\cite{kipf2017semi}. GCN's capability of leveraging both information of nodes (representing features of entities) and edges (representing connections or relationships between nodes), allowing for feature aggregation in the network level. Advanced versions of GCN include graph attention networks (GATs) ~\cite{Veli}, graph transformer networks ~\cite{yun2019graph} etc. Many research studies have employed graph-based methods to predict chronic diseases~\cite{ochoa2022graph}, mental disorders~\cite{qin2022using,wen2023prediction}, and Alzheimer's disease~\cite{zhu2022interpretable}. In this paper, we propose a multimodal framework integrating both image features and clinical features by building a contrastive graph cross-view learning approach where the graph represents the similarity of individuals in the embedded space for detecting the PD. 

\begin{figure*}
\centering
\includegraphics[width=0.75\textwidth]{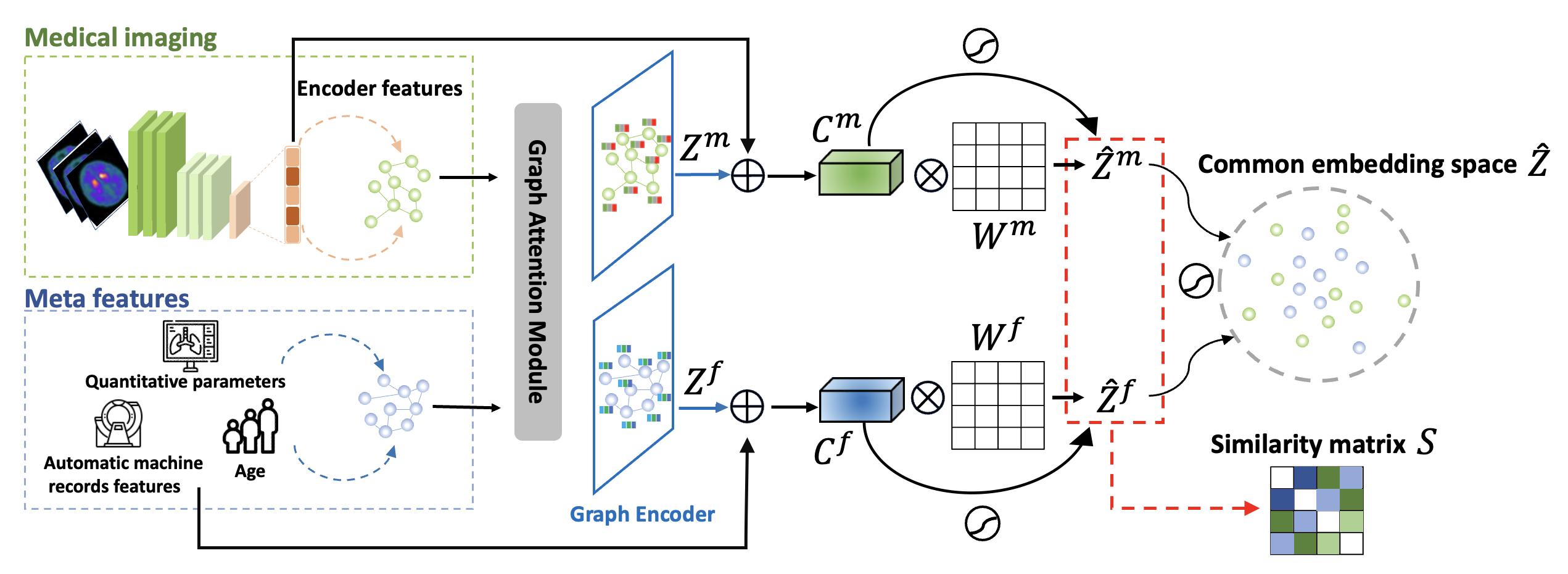}
\caption{The workflow of multimodal contrastive cross-view graph learning framework.}
\label{fig:Fig1}
\end{figure*}

\section{RELATED WORK}

\subsection{Multi-Modality Fusion Methods}
Image-based models historically relied on unimodal input, which were commonly used for disease classification. However, these structured medical images alone are not effective integrating the physiological or numerical characteristics of patients. Existing work showed the effectiveness of integrating brain Single-photon Emission Computed Tomography (SPECT) images and DNA methylation data in a multi-co-attention model~\cite{Taylor_2019}, and also hybrid models that integrate CNN and LSTM are used to incorporate dynamic and static speech features to diagnose early physically incapacitating symptoms in the PD patients~\cite{Lilhore_2023}. 

\subsection{Graph Representation Learning}
Single graph structure is proposed as a way to combine image and non-image characteristics ~\cite{Chen_2021,Huang_2022}. However, the image and non-image data belong to different types or require two types of graph network structures, such as topology graph and feature graph, should be considered for better feature fusion ~\cite{Wang_2020}. To improve multimodal classification performance, the GAT incorporates a mechanism of attention to build a graph containing image features as well as clinical features, with category-wise attention and updated node features incorporating importance scores ~\cite{Cui_2021}.

\subsection{Supervised Graph Contrastive Learning}
Supervised contrastive learning, as demonstrated by its significant successes in image applications such as visual representations ~\cite{Khosla_2020,SimSLR}, primarily aims to enhance the similarity among positive pairs while simultaneously augmenting the dissimilarity between negative pairs ~\cite{SCGC}. Contrastive learning with two graph views was proven effective in fMRI-based neuroimaging classification in a previous medical study that sought to improve the diagnosis of neurological disorders in autism and dementia ~\cite{Peng_2022_GATE}. Investigations have been conducted on to the design of contrastive losses, such as the InfoNCE loss ~\cite{Oord_2018}, which maximizes the consistency of positive pairs and uses negative sampling to increase the number of negative pairs from different batches for $k$ classes.

\section{METHODOLOGY}\label{sec2}

\subsection{Features Extraction and Graph Construction }

We can denote a dataset of $N$ patients with the $i$-th patient's SPECT image denoted as $X^{m}_{i}$ and non-image features denoted as $X^{f}_{i}$, where $X^{m}_{i} \in \mathbb{R}^{\mu \times \nu}$ and $X^{f}_{i} \in \mathbb{R}^{F}$ with $F$ features, and the label matrix $Y \in \mathbb{R}^{N \times C}$ with $C$ classes. The multimodal dataset can be described as $\{ X^{m}_{i}, X^{f}_{i}, Y_{i} \}^{N}_{i=1}$, where $Y_i$ is the $i$-th row of $Y$. In the first stage, we use a CNN-based autoencoder $h(\cdot)$ for image feature extraction and flatten the output image feature matrix $Q^{m}$ in the final layer. Then, we constructed two adjacency matrices $A^{m}$, $A^{f} \in \mathbb{R}^{N \times N}$  using the $K$ nearest neighbors (KNN) algorithm based on the features obtained from the CNN model encoder and the non-image clinical features (such as patient age, biomarkers, symptoms, etc.) for each images encoder and clinical features subject. Our proposed framework considers two graphs, $\mathcal{G}^{m}(X^{m}, A^{m})$ and $\mathcal{G}^{f}(X^{f}, A^{f})$ , as different domain inputs. We also considered a self-loop adjacency matrix $\hat{A}=A+I$ between the patients. In a K-neighborhood, two data points $i$ and $j$ are connected by an edge $e_{(i,j)}$ if $i$ is amongst the $K$ nearest neighbors of $j$, or vice versa.

\subsection{Graph Encoder and Cross-View Fusion}

In this stage, we constructed two GATs to learn the graph structures of $\mathcal{G}^{m}(A^{m}, X^{m})$ and $\mathcal{G}^{f}(A^{f}, X^{f})$ as depicted in (Fig. \ref{fig:Fig1}). We utilize encoded image features $X^{m}$ and clinical features $X^{f}$ as node attributes for the GAT inputs. Moreover, we introduced a GAT architecture that incorporates dual perspectives, enabling the generation of embeddings for neighboring nodes. The most common expression for attention coefficients as applied to our two cross-views is as follows:

\begin{equation}
\alpha_{ij}=\frac{exp\left(LeakyReLU(\vec{a}^T_{ij}\left[W\vec{h}_{i}  \mathbin\Vert W\vec{h} _{j} \right] \right)}{\sum^{N}_{k=1}exp\left(LeakyReLU(\vec{a}^T_{ij}\left[W\vec{h}_{i}  \mathbin\Vert W\vec{h}_{k} \right] \right)}
\end{equation}

Finally, we can obtain each feature $\vec{h}^{\prime}$ for two cross-views feature representations as shown in Equation (2):

\begin{equation}
\vec{h}^{\prime} =   \sigma \left( \frac{1}{K} \sum^{N}_{k=1}\sum_{j\in \mathcal{N}_{i}}\alpha^{k}_{ij}W^{k}\vec{h}_{j} \right),
\end{equation}
where $\alpha^{k}_{ij}$ and $W^{k}$ are the attention mechanism and linear transformation weight matrix, and $\mathcal{N}_{i} $ denotes the set of neighborhood nodes of $i$.

The extracted nodes representation from the GAT output is denoted as $Z^{m} = f(\mathcal{G}^{m})$, and the non-image feature embeddings are represented by \(Z^{f} = f(\mathcal{G}^{f})\), where $Z^{m}$ and $Z^{f}$ are the embeddings in a low-dimensional space $\mathbb{R}^{F^{\prime}}$. Afterward, we concatenated the encoder matrix $Q^{m}$ with $Z^{m}$ to form $C^{m}$, and the clinical features $X^{f}$ with $Z^{f}$ to create $C^{f}$ as shown below:

\begin{equation}
C^{m} = \left[Q^{m}  \mathbin\Vert Z^{m} \right]
\end{equation}
\begin{equation}
C^{f} = \left[  X^{f} \mathbin\Vert Z^{f}\right]
\end{equation}

\noindent where the $C^{m} \in \mathbb{R}^{N \times (F+F^\prime)}$ and $C^{f}\in \mathbb{R}^{N \times (F+F^\prime})$  represent the two concatenated matrices from cross-views. The improved fusion embedding $\hat{Z}^{m}$ and $\hat{Z}^{f}$, can be obtained by $\sigma( C^{m}W^{m})$ and $\sigma( C^{f}W^{f})$ respectively, where $W^{m}$ and $W^{f}$ are trainable weight matrices, and $\sigma(\cdot)$ is non-linear activation function.

\subsection{Contrastive Cross-View Loss}

In order to learn the common embedding $\hat{Z}$, we fuse the two cross-views of node embeddings between $\hat{Z}^{m}$ and $\hat{Z}^{f}$ as

\begin{equation}
\hat{Z}= \hat{Z}^{m} + \hat{Z}^{f}
\end{equation}

To better integrate the feature spaces of image and non-image data in the same embedded space, we constructed a similarity matrix $S  \in \mathbb{R}^{N \times N}$ for each pair of similar patients using the final embedding $\hat{Z}$ learned from the model. We can define the similarity between the $i$-th and $j$-th patients as follows:

\begin{equation}
S_{ij} =  \hat{Z}_{i} \cdot (\hat{Z}_{j})^{T}, \forall i,j \in \left [ 1,N \right ]
\end{equation}
In order to enhance the effectiveness of fusing two types of view embeddings in contrastive learning, we have designed positive and negative losses to capture the differences in distance between positive and negative pairs in terms of the similarity and dissimilarity of our samples. The definitions of positive pair $D_{pos} = S \odot (\hat{A}^{m} \odot \hat{A}^{f}) $, while negative pair $D_{neg} = (\mathbb{I}-S) \odot \left[(\mathbb{I}  - \hat{A}^{m}) \odot (\mathbb{I}  - \hat{A}^{f}) \right]$, where $\mathbb{I}$ denotes the matrix with all elements being 1 with the related dimension, and the two adjacency matrices with self-looped is denoted as $\hat{A}^{m}$ and $\hat{A}^{f}$ ~\cite{Liu_2022}. Then, we can present the loss function of positive and negative pairs as shown below:
\begin{equation}
\left\{\begin{aligned}
\mathcal{L}_{pos} &= -  \lVert D_{pos} \cdot Y\lVert^{2}_{2} \\
\mathcal{L}_{neg}&=  - \lVert   \max\{ D_{neg} - \delta\mathbb{I}, 0\} (\mathbb{I}-Y) \lVert^{2}_{2}    \\
\end{aligned}
\right.
\label{eq:pos_neg_los}
\end{equation}
\noindent where the $\delta > 0$ is the controllable margin and $Y$ is the label matrix. By using Eq.~\ref{eq:pos_neg_los}, we can ultimately obtain the combined losses, incorporating both positive and negative loss, written as: $\mathcal{L}_{contrastive} = \mathcal{L}_{pos} + \mathcal{L}_{neg}$. By minimizing $\mathcal{L}_{contrastive}$, the similarity intra-class and the dissimilarity inter-class can be maximized. 
\subsection{Optimization Objective Function} 
To optimize the loss function and predict final disease probability, we considered embedding both $\hat{Z}^{m}$ and $\hat{Z}^{f}$ in the supervised classification loss using the softmax function. The cross-entropy loss function can be written as:

\begin{equation}
\mathcal{L}_{m} = -\sum_{i=1}^{N}y_{i}^{T} \rm ln(softmax(\hat{y}_{i}^{m}))
\end{equation}

\begin{equation}
\mathcal{L}_{f} = -\sum_{j=1}^{N}y_{j}^{T} \rm ln(softmax(\hat{y}_{j}^{f}))
\end{equation}

During the optimization process, we also designed the overall loss function to combine cross-entropy and contrastive loss from the two cross views. To effectively improve the cross-graph view module, we also took into account the mean square error loss between the similarity matrix $S$ and the diagonal matrix $D_{ii}=\sum_{i}A_{ii}$ when computing the clustering of the view structure of the two modules.

\begin{equation}
\mathcal{L}_{diag} = \frac{1}{N}\sum^{N}_{i,j} (S_{ij} - D_{ii})^{2}
\end{equation}

We use the $\beta$ coefficient to control the optimization weight of the overall loss defined as follows:
\begin{equation}
\mathcal{L} = (1- \beta)(\mathcal{L}_{m}  + \mathcal{L}_{f}) + \beta\mathcal{L}_{contrastive} + \mathcal{L}_{diag} 
\end{equation}

where $\beta$ can be set between 0 and 1. The contribution level of different losses is controlled through the coefficient of $\beta$.

\section{EXPERIMENTS}

\subsection{Dataset}
Our data was collected at Kaohsiung Chang Gung Memorial Hospital in Taiwan from January 2017 to Jun 2019 with 416 patients~\cite{Ding_2021}. The data was annotated by four expert physicians  to provide a label either as normal or abnormal PD. Tc99m TRODAT SPECT/CT images were acquired using a hybrid SPECT/CT system (Symbia T, Siemens Medical Solution). SPECT images were obtained with 30s per step, acquiring 120 projections over a circular 360-degree rotation using low-energy, high-resolution parallel-hole collimators.

\begin{figure*}[t]
\centering
\includegraphics[width=0.95\textwidth,height=0.34\textwidth]{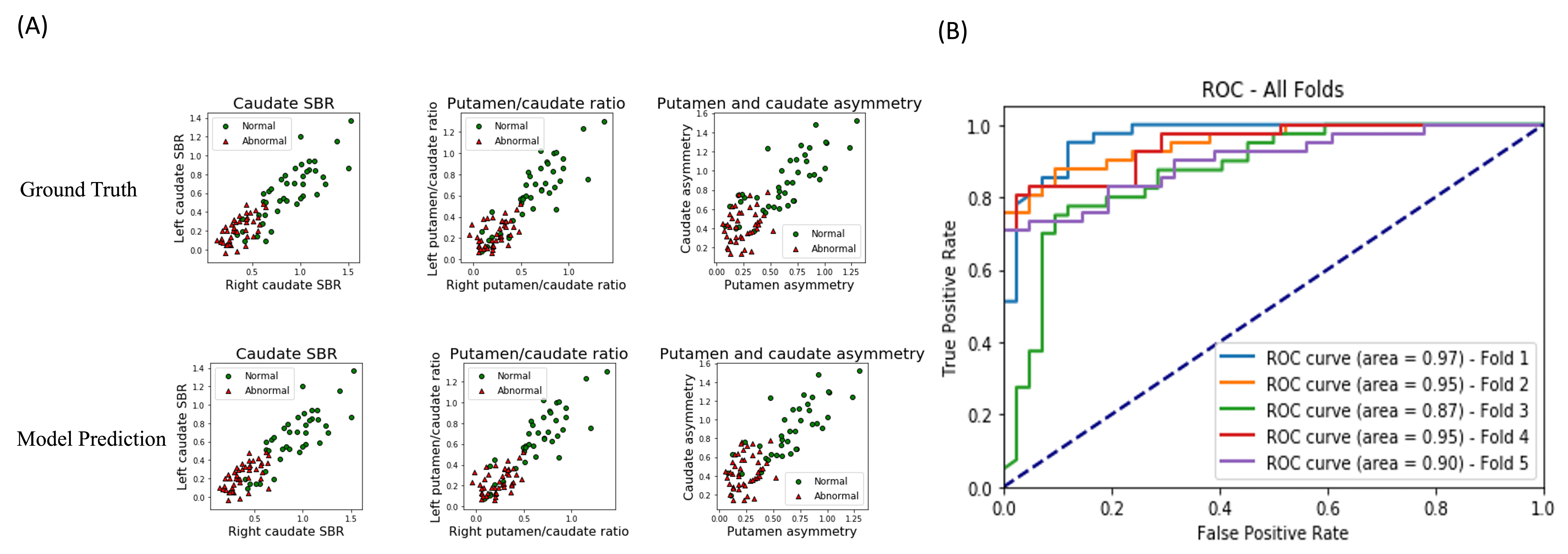}
\caption{Visualization of the predictions of the two-graph cross-view GAT model, incorporating three variables from twelve parameters. Figure (A) Scatter plots of three parameters derived from DaTQUANT to explore the data distribution for normal versus abnormal TRODAT SPECT images. Figure (B) Five-fold cross-validation of ROC curves for each testing set.}
\label{fig:Fig2}
\end{figure*}

\vspace{-2mm}

\begin{table}[h]
\centering
\begin{tabular}{@{}lcc@{}}
\toprule
                      & \textbf{Normal} & \textbf{Abnormal} \\ \midrule
No. of Patients       & 208             & 204               \\
\hline
S-R                   & 0.74    ±0.23      & 0.40  ±0.18           \\
S-L                   & 0.75      ± 0.23      & 0.41    ± 0.17        \\
AP-R                  & 0.76     ± 0.25       & 0.38   ± 0.18           \\
AP-L                  & 0.78     ± 0.25     & 0.40   ± 0.17           \\
PP-R                  & 0.55    ± 0.25       & 0.25    ± 0.16          \\
PP-L                  & 0.54   ± 0.25         & 0.25   ± 0.15           \\
C-R                   & 0.79    ± 0.24        & 0.49    ± 0.23          \\
C-L                   & 0.81   ± 0.24         & 0.51    ± 0.23          \\
P/C-R                 & 0.96   ± 0.09         & 0.91     ± 0.09         \\
P/C-L                 & 0.95     ± 0.08       & 0.91     ± 0.09         \\
PA                    & 0.04    ± 0.03        & 0.05     ± 0.05         \\
CA                    & 0.05    ± 0.04        & 0.06     ±0.05         \\ \bottomrule
\end{tabular}
\caption{Summary statistics for 12 non-image automatic semi-quantification features were generated using the DaTQUANT software.}
\label{tab:my_label}
\end{table}

After reconstruction with CT attenuation correction, SPECT images were imported into DaTQUANT for automatic semi-quantification ~\cite{Brogley_2019}. Twelve parameters were obtained from DaTQUANT: Striatum Right (S-R), Striatum Left (S-L), Anterior Putamen Right (AP-R), Anterior Putamen Left (AP-L), Posterior Putamen Right (PP-R), Posterior Putamen Left (PP-L), Caudate Right (C-R), Caudate Left (C-L), Putamen/Caudate Ratio Right (P/C-R), Putamen/Caudate Ratio Left (P/C-L), Putamen Asymmetry (PA), and Caudate Asymmetry (CA) as shown in Table 1. Afterwards, images from four patients were removed from the dataset due to quality issues. We used the remaining  412 SPECT images and quantitative DaTQUANT data for model training (n=312) and testing (n=100), and we conducted five-fold cross validation. 
\vspace{-3mm}
\subsection{Ablation Study}
\vspace{-3mm}
\begin{table}[h]
\vspace{-3mm}
\caption{\footnotesize Performance comparison between the proposed model and machine learning methods using image and non-image data.}
\renewcommand{\arraystretch}{1}
\resizebox{\columnwidth}{!}{%
\Huge
\begin{tabular}{c|c|c|c|c|lcc}
\hline
                                 & Model                        & Backbone                     & Image                 & Non-image             &       & Normal     & Abnormal    \\ \hline
\multirow{16}{*}{Baseline}       & \multirow{4}{*}{Logistic}    & \multirow{4}{*}{-}           & \multirow{8}{*}{No}   & \multirow{8}{*}{Yes}  & ACC  & \multicolumn{2}{c}{0.80}  \\
                                 &                              &                              &                       &                       & F1    & 0.82       & 0.79        \\
                                 &                              &                              &                       &                       & SEN & 0.82       & 0.79        \\
                                 &                              &                              &                       &                       & PRE & 0.82       & 0.79        \\ \cline{2-3} \cline{6-8} 
                                 & \multirow{4}{*}{Xgboost}     & \multirow{4}{*}{-}           &                       &                       & ACC  & \multicolumn{2}{c}{0.79} \\
                                 &                              &                              &                       &                       & F1    & 0.80        & 0.76        \\
                                 &                              &                              &                       &                       & SEN & 0.82       & 0.75        \\
                                 &                              &                              &                       &                       & PRE & 0.79       & 0.78        \\ \cline{2-8} 
                                 & \multirow{4}{*}{ResNet18}    & \multirow{4}{*}{-}           & \multirow{24}{*}{Yes} & \multirow{8}{*}{No}   & ACC  & \multicolumn{2}{c}{0.89} \\
                                 &                              &                              &                       &                       & F1    & 0.90        & 0.88        \\
                                 &                              &                              &                       &                       & SEN & 0.92       & 0.87        \\
                                 &                              &                              &                       &                       & PRE & 0.89       & 0.90         \\ \cline{2-3} \cline{6-8} 
                                 & \multirow{4}{*}{2-layer CNN} & \multirow{4}{*}{-}           &                       &                       & ACC  & \multicolumn{2}{c}{0.85} \\
                                 &                              &                              &                       &                       & F1    & 0.85       & 0.84        \\
                                 &                              &                              &                       &                       & SEN & 0.83       & 0.87        \\
                                 &                              &                              &                       &                       & PRE & 0.88       & 0.82        \\ \cline{1-3} \cline{5-8} 
\multirow{4}{*}{Existing Methods}                                 & \multirow{4}{*}{AM-GCN}  & \multirow{4}{*}{GCN+Attention}           &                       &                       & ACC  & \multicolumn{2}{c}{0.86} \\
                                 &                              &                              &                       &                       & F1    &    0.86    &     0.85    \\
                                 &                             &                              &                       &                       & SEN &    0.83    &     0.88    \\
                                 &                              &                              &                       &                       & PRE &    0.89    &    0.82     \\ \cline{1-3} \cline{6-8} 
\multirow{16}{*}{Proposed Model} & \multirow{8}{*}{GCN+GCN}     & \multirow{4}{*}{2-layer CNN} &                       & \multirow{16}{*}{Yes} & ACC  & \multicolumn{2}{c}{0.87} \\
                                 &                              &                              &                       &                       & F1    & 0.88       & 0.85        \\
                                 &                              &                              &                       &                       & SEN & 0.88       & 0.85        \\
                                 &                              &                              &                       &                       & PRE & 0.87       & 0.86        \\ \cline{3-3} \cline{6-8} 
                                 &                              & \multirow{4}{*}{ResNet18}    &                       &                       & ACC  & \multicolumn{2}{c}{0.88} \\
                                 &                              &                              &                       &                       & F1    & 0.89       & 0.88        \\
                                 &                              &                              &                       &                       & SEN & 0.87       & 0.90         \\
                                 &                              &                              &                       &                       & PRE & 0.91       & 0.85        \\ \cline{2-3} \cline{6-8} 
                                 & \multirow{8}{*}{GAT+GAT}     & \multirow{4}{*}{2-layer CNN} &                       &                       & ACC  & \multicolumn{2}{c}{0.86} \\
                                 &                              &                              &                       &                       & F1    & 0.87       & 0.84        \\
                                 &                              &                              &                       &                       & SEN & 0.88       & 0.83        \\
                                 &                              &                              &                       &                       & PRE & 0.85       & 0.86        \\ \cline{3-3} \cline{6-8} 
                                 &                              & \multirow{4}{*}{ResNet18}    &                       &                       & ACC  & \multicolumn{2}{c}{\textbf{0.91}} \\
                                 &                              &                              &                       &                       & F1    & \textbf{0.92}       & \textbf{0.90}       \\
                                 &                              &                              &                       &                       & SEN & \textbf{0.92}       & \textbf{0.90}        \\
                                 &                              &                              &                       &                       & PRE & \textbf{0.92}       & \textbf{0.90}        \\ \hline
\end{tabular}
}
\label{table:example}
\end{table}

\begin{figure}
\centering
\includegraphics[width=0.3\textwidth]{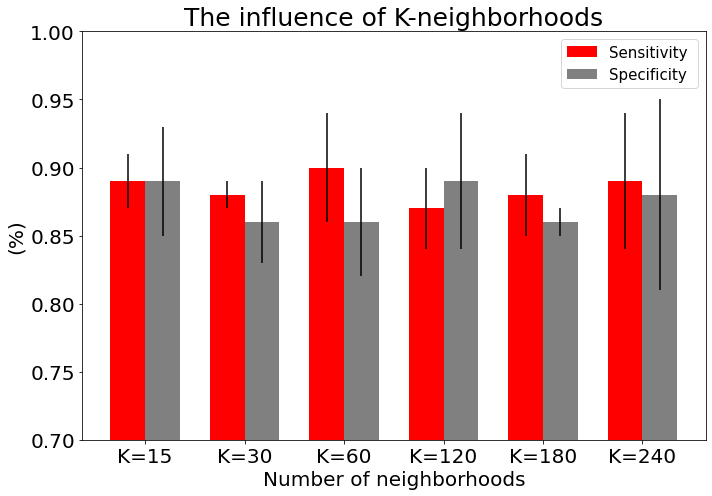}
\caption{The mean and standard deviation performance of our proposed model in terms of sensitivity and specificity across five runs on testing data, based on a varying number of K-neighborhoods.}
\label{fig:Fig3}
\end{figure}
In this study, we conducted a comparative analysis using popular machine learning algorithms (logistic regression and Xgboost)  as the baseline methods ~\cite{Hsu_2019} and existing methods AM-GCN ~\cite{wang2020gcn} in the ablation experiments using non-image DaTQUANT data. Additionally, we employed a two-layered CNN model with a ResNet18 backbone, specifically utilizing ResNet18 for the classification of SPECT images. As shown in Table 2, relying solely on the CNN model for prediction did not lead to superior performance. 

In our research, we began  by extracting image features using a CNN model and non-imaging DaTQUANT variables, followed by constructing two cross-views of the graph representation. We subsequently concatenated these different modalities to improve the predictive capability after model fusion. Table 2 showcases our experimental results from utilizing two methods (GCN and GAT) for learning graph structures and generating embeddings. The results revealed that, when a cross-view approach was employed, the GAT method achieved a macro average accuracy rate of 0.91, along with F1, sensitivity, and precision scores of 0.92 in normal and abnormal classes. Additionally, our method can achieve an 5-fold cross-validated AUC of 0.93  as shown in (Fig. \ref{fig:Fig2} (B)).

To conclude, we investigated the sensitivity of the overall model to parameter $K$ in our proposed method for constructing $K$-nearest neighbor graphs. Figure \ref{fig:Fig3} clearly demonstrates the results of five experimental runs, showcasing a remarkably robust performance with an average sensitivity and specificity of 0.89.

\section{RESULTS AND DISCUSSION}
In summary, we successfully integrated the automatic semi-quantification features from both image and non-image data to enhance prediction accuracy in the PD classification task. By utilizing cross-view graph structured information, we successfully predicted the distribution of twelve non-imaging parameters for both normal and abnormal cases within a low-dimensional space, resulting in effective clustering and interpretation, as depicted in (Fig. \ref{fig:Fig2} (A)). Our research findings indicate that models based solely on CNNs can have certain limitations in interpreting image features. However, those limitations can be overcome to some extent through the integration of non-imaging data and the application of contrastive loss learning, which significantly enhanced the overall performance and predictive capacity of our model.

\section{Acknowledgment:} We are grateful to the Department of Nuclear Medicine at Kaohsiung Chang Gung Memorial Hospital for providing us with comprehensive data and data labeling support. Research reported in this publication was partially supported by the National Institute of Biomedical Imaging and Bioengineering of the National Institutes of Health under Award Number R21EB033455  and research grant from New Jersey Health Foundation under grant number PC 40-23. The content is solely the responsibility of the authors and does not necessarily represent the official views of the National Institutes of Health nor the New Jersey Health Foundation.

\bibliographystyle{IEEEbib}
\bibliography{strings,refs}
\end{document}